\documentclass[letterpaper]{article} % DO NOT CHANGE THIS
\usepackage{aaai25}  % DO NOT CHANGE THIS
\usepackage{times}  % DO NOT CHANGE THIS
\usepackage{helvet}  % DO NOT CHANGE THIS
\usepackage{courier}  % DO NOT CHANGE THIS
\usepackage[hyphens]{url}  % DO NOT CHANGE THIS
\usepackage{graphicx} % DO NOT CHANGE THIS
\usepackage{natbib}  % DO NOT CHANGE THIS AND DO NOT ADD ANY OPTIONS TO IT
\usepackage{caption} % DO NOT CHANGE THIS AND DO NOT ADD ANY OPTIONS TO IT
\usepackage{algorithm}
\usepackage{algorithmic}
\usepackage{booktabs}
\usepackage{multirow}
\usepackage{newfloat}
\usepackage{listings}
\usepackage{subfig}
\DeclareCaptionStyle{ruled}
{labelfont=normalfont,labelsep=colon,strut=off} % DO NOT CHANGE THIS
\floatstyle{ruled}
\newfloat{listing}{tb}{lst}{}
\floatname{listing}{Listing}
\title{Don’t Just Demo, Teach Me the Principles: A Principle-Based Multi-Agent Prompting Strategy for Text Classification}
\author {
    Peipei Wei,
    Dimitris Dimitriadis,
    Yan Xu,
    Mingwei Shen
}
\affiliations {
    Amazon\\
    \{peipeiw, dbdim, yanxuml, mingweis\}@amazon.com
}

\usepackage{bibentry}
\begin{document}

\maketitle

\begin{abstract}
We present PRINCIPLE-BASED PROMPTING, a simple but effective multi-agent prompting strategy for text classification. It first asks multiple LLM agents to independently generate candidate principles based on analysis of demonstration samples with or without labels, consolidates them into final principles via a finalizer agent, and then sends them to a classifier agent to perform downstream classification tasks. Extensive experiments on binary and multi-class classification datasets with different sizes of LLMs show that our approach not only achieves substantial performance gains (1.55\% - 19.37\%) over zero-shot prompting on macro-F1 score but also outperforms other strong baselines (CoT and stepback prompting). Principles generated by our approach help LLMs perform better on classification tasks than human-crafted principles on two private datasets. Our multi-agent PRINCIPLE-BASED PROMPTING approach also shows on-par or better performance compared to demonstration-based few-shot prompting approaches, yet with substantially lower inference costs. Ablation studies show that label information and the multi-agent cooperative LLM framework play an important role in generating high-quality principles to facilitate downstream classification tasks.
\end{abstract}


\section{Introduction}
In recent years, transformer-based language models with attention mechanisms have deeply revolutionized the field of NLP. Particularly, decoder-only transformer language models, such as GPT-series models, demonstrate impressive emerging capabilities after scaling up the pre-training corpora and model sizes—capabilities not seen in their smaller predecessors such as BERT-based models \cite{zheng2023take}. One of these capabilities is In-Context Learning (ICL) \cite{brown2020language}. Equipped with knowledge acquired during the pre-training stage, these large language models (LLMs) are able to perform various tasks with only task instructions and a few demonstrations, without any parameter updates. Despite their surprisingly good zero-shot and few-shot performance on a wide range of tasks such as general QA, reasoning, and text generation, their performance still significantly lags behind fine-tuned models for text classification \cite{sun2023text}.

On the other hand, these fine-tuned models heavily depend on human annotations, which are not only costly and time-consuming but also sometimes unavailable. Accordingly, leveraging zero-shot or few-shot ICL capabilities of LLMs for text classification has become an important research topic. However, ICL relies on prompt engineering and human expertise in designing demonstration questions, intermediate reasoning steps, and final answers for LLMs to generalize to a variety of unseen queries. Additionally, increasing the number of demonstrations in few-shot settings leads to increased inference costs and may exceed the maximum input length imposed by LLMs.

When humans work on complicated tasks, they usually follow Standard Operating Procedures (SOPs) to ensure that anyone with varying degrees of domain and task-specific knowledge can perform the task with consistently high quality. These SOPs are written by domain experts who have gained expertise by analyzing numerous concrete examples and extracting common principles from them. Inspired by this, we ask: can we mimic the same procedure to generate task-specific principles based on analysis of a handful of demonstrations and then feed them back to LLMs to help mitigate the limitation of lack of task-specific knowledge in ICL?

Previous studies show that adding complex class descriptions as additional inputs to a pre-trained transformer backbone via cross-encoder architecture can significantly boost classification performance under zero-shot and few-shot settings \cite{de2023semantic}. Intuitively, injecting more knowledge-intensive principles should also help improve LLMs' ICL performance.

In this paper, we present PRINCIPLE-BASED PROMPTING for zero-shot text classification. It utilizes a multi-agent collaboration framework to auto-generate principles for each classification task. First, it employs multiple LLM agents to generate candidate principles from demonstrations with or without labels. In the prompts, it explicitly instructs LLMs to extract key principles that can distinguish each class based on analysis of provided demonstrations. Then, all LLM agents send their principle candidates to a central agent for finalization, which selects the best principles for downstream classification tasks. Our approach demonstrates substantial performance gains over other strong baseline ICL approaches, such as Chain-of-Thought (CoT) \cite{wei2022chain} and step-back prompting \cite{zheng2023take}, in zero-shot ICL settings. The performance is also very competitive compared to few-shot ICL. In summary, our contributions are as follows:
\begin{itemize}

\item We conduct extensive experiments on three public and two private datasets with two LLMs (flan-t5-xxl and flan-ul2) and show that our approach substantially boosts zero-shot ICL performance on both binary and multi-class classification problems over vanilla prompting as well as strong ICL baselines. We also show on-par or even better classification performance using automatically generated SOPs compared to human-generated SOPs on two private datasets.
\item Our approach demonstrates competitive performance even compared to few-shot ICL. Unlike previous work, our multi-agent approach boosts performance while requiring much shorter input token lengths, resulting in significantly reduced inference costs.
\item Our PRINCIPLE-BASED PROMPTING approach significantly outperforms fine-tuned RoBERTa-large under low-resource settings, although the performance of supervised models tends to improve when more labeled data becomes available.
\item Through ablation studies, we have identified that label information and the reasoning capabilities of LLMs are key contributors to extracting high-quality principles for downstream classification tasks. We demonstrate the advantages of a multi-agent approach over a single-agent approach. Additionally, we show that selecting more capable LLMs to generate candidate principles and focusing on collaboration rather than competition among LLM agents are important factors when constructing a multi-agent LLM collaboration framework for text classification.
\end{itemize}

\section{Related work} 
\subsubsection{Demonstration and label relationship}  
Supervised ML models rely heavily on drawing mappings between representations of training examples and their label information to make predictions on unseen examples. Surprisingly, early research on ICL shows that ground truth in demonstration-label mapping is not as important, as showing demonstrations with random labels only leads to minimal performance drops on a range of classification tasks \cite{min2022rethinking}. However, later research points out the limitations of this study and arrives at a different conclusion: the correct correspondence between examples and labels is essential to ensure ICL performance \cite{kossen2023context}. The previous biased conclusion could be attributed to the use of binary (accuracy) instead of probabilistic metrics, relatively weaker LLMs that are mostly under 20B parameters, and focus on only one few-shot setting (16 demos). Thus, although LLMs predominantly rely on knowledge acquired during pre-training to perform downstream tasks, they indeed can learn new tasks from in-context information, which motivates this work to find an alternative approach to providing more effective context information for LLM ICL than the commonly used demonstration-based approach. In our experiments, we also conduct ablation studies to explore the importance of label information on the quality of principles generated.
\subsubsection{Number of demonstrations}
Supervised ML algorithms are data-hungry and require a substantial amount of labeled training data to ensure model performance. Under ICL few-shot settings, previous work shows that adding more than one demonstration might not be necessary due to only marginal performance improvements \cite{chen2023many}. As Chen suggests, this indicates that the use of demonstrations is inefficient and the information provided by randomly selected demonstrations is most likely redundant. In some cases, multiple demonstrations can even hurt performance due to misguidance or negative interference among them \cite{chen2023many}. This leads to our research question: under the same input length constraint, can we design more concise but knowledge-intensive contexts as alternatives to few-shot demonstrations to better guide LLMs in performing downstream classification tasks? We also conduct ablation studies to explore the importance of the number of demonstrations on the quality of principles generated.

\subsubsection{Single-Agent vs. Multi-Agent LLM Framework}
Text classification, as one of the most fundamental NLP tasks, appears to be straightforward in the sense that LLMs only need to output one or more class labels from a predefined label space. However, it can actually be quite complicated and even more challenging due to the implicit nature of the reasoning process in comparison to other tasks. Most research on LLM ICL attempts to enhance model performance either by decomposing complex tasks into multiple steps or by providing LLMs with relevant domain- and task-specific data as additional context, such as the Retrieval Augmented Generation (RAG) approach. For instance, Chain-of-Thought (CoT) prompting first prompts the LLM to break problem-solving into multiple steps and then derives the final answer by following a step-by-step thought process \cite{wei2022chain}. Focusing on QA questions, stepback prompting \cite{zheng2023take} runs inference on the same LLM twice by first asking LLMs to provide abstract principles or concepts to help resolve the original question before answering it. To improve LLMs' performance on text classification, for each data point, Clue And Reasoning Prompting (CARP) \cite{sun2023text} includes multiple steps in a single prompt by asking the same LLM to first find superficial clues (e.g., keywords, tones, semantic relations, references, etc.) based on which final decisions are made after reasoning steps. CARP also leverages knowledge acquired through supervised fine-tuning on labeled datasets to search for more effective demonstrations for ICL. 

Recently, the multi-agent framework has gained popularity and has been shown to greatly improve LLMs' performance on complicated tasks such as long-context QA, multi-hop QA, math, and reasoning \cite{shridhar2022distilling, wang2022self}. For instance, the multi-agent debate framework can improve LLMs' reasoning capability, factuality, and inter-consistency in mathematical and multiple-choice commonsense reasoning tasks, as well as output quality in open-ended generation tasks, in comparison to their single-agent counterparts \cite{du2023improving, xiong2023examining, chan2023chateval}. In our multi-agent implementation of the principle-based approach, we try competitive and collaborative paradigms and evaluate their effectiveness.

Performance improvements provided by single- or multi-agent solutions mentioned above, using either self-ensemble (multiple inferences on the same LLM agent) or heterogeneous ensemble (multiple inferences on different LLM agents) approaches, usually come at significantly increased inference costs due to multiple LLM inferences and/or communication costs across different agents. Our research question is: can we achieve the same performance improvement without significantly increasing inference costs for text classification? Unlike other tasks such as long-context QA or text generation tasks, the label space for most text classification tasks is finite and relatively limited. Thus, the search space for an optimal principle should also be bounded. Accordingly, we propose to implement an effective and efficient multi-agent LLM framework to auto-generate a single all-inclusive SOP for each task and reuse it for inference on all data points. We believe that, in addition to improving performance, the shared principle can also help ensure consistency in classification predictions.

\section{Methods} 
\begin{figure*}[htbp]
    \centering
    \subfloat{
    \includegraphics[width=0.5\textwidth]{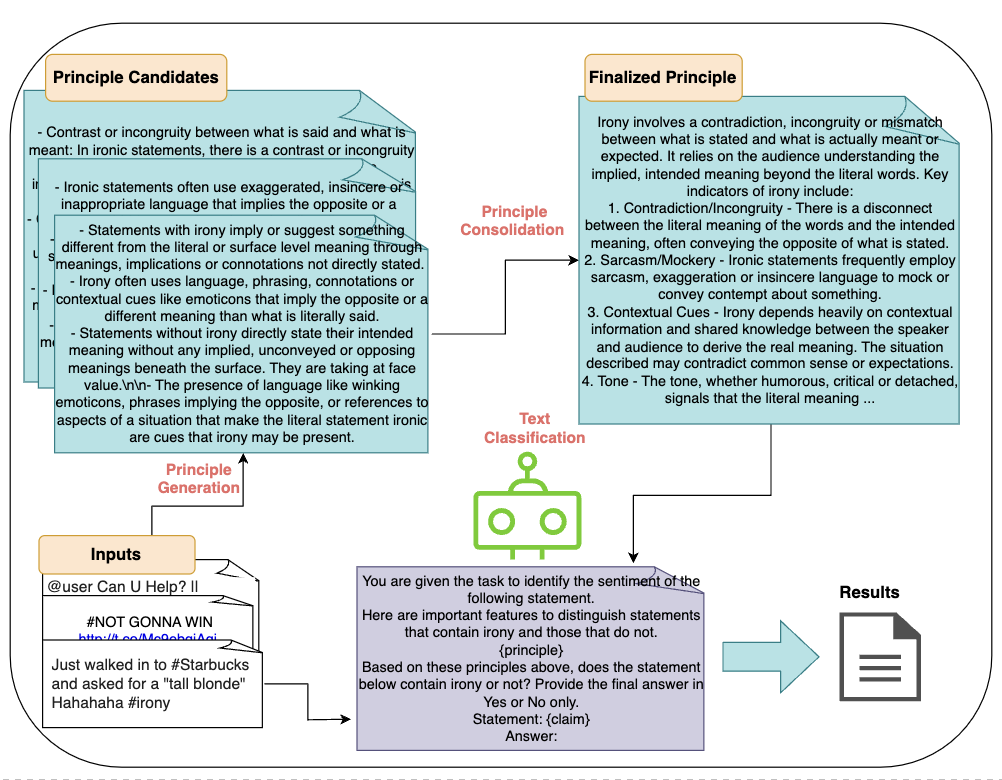}
        \label{fig:pipeline}
    }
    \hfill
    \subfloat{
        \includegraphics[width=0.4\textwidth]{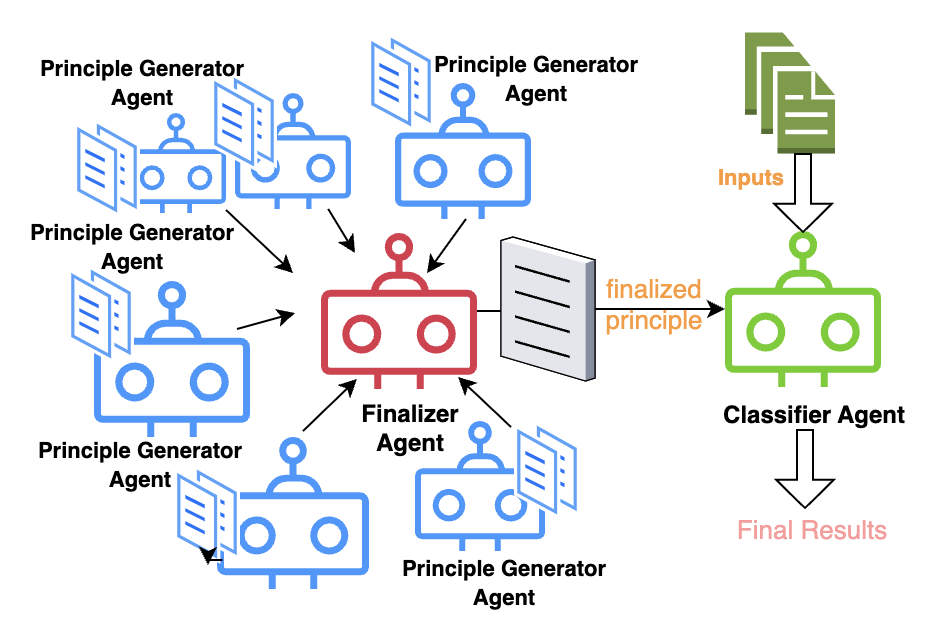}
        \label{fig:multiagents}
    }
    \caption{Pipeline and Multiagent illustrations of PRINCIPLE-BASED PROMPTING}
    \label{fig:multiagents1}
\end{figure*}

\textbf{PRINCIPLE-BASED PROMPTING} is motivated by the observation that when performing classification tasks, human beings usually start to build their mental models after reviewing a few concrete examples by summarizing common key principles. Humans tend to rely on abstracted principles since we have limited memory capacity to remember overwhelmingly large amounts of detailed data points. The more comprehensive these principles are to include different scenarios, the more helpful they should be for performing the same task on unseen data. As we see later that in our two internal datasets (Product Classification 1 and Product Classification 2, PC1 and PC2), we have principles manually drafted by domain experts for each task to help ensure annotation quality. In the experiments section, we also investigate whether text classification via ICL with principles generated by our multi-agent framework can outperform their human-generated counterparts. We implement our PRINCIPLE-BASED PROMPTING strategy via a multi-agent LLM framework. It consists of three major steps, each of which can be completed by one or multiple LLM agents (see Figure \ref{fig:multiagents1} ).

\subsubsection{Principle Generation}
Before tackling the classification problem, we first ask the multi-agent LLMs to analyze a few randomly sampled demonstrations with or without label information on their own. Then, we ask them to generate principles to distinguish each class based on their analysis. Since principles are generated at the task level, additional inference costs only occur for each principle generated, which is almost negligible in comparison to the inference costs for entire datasets.

In this step, we experiment with a diverse set of six different LLMs, ranging from open to closed models in various model sizes: two open-source LLMs from Huggingface: FLAN-T5-XX \cite{chung2024scaling} with 11B parameters and FLAN-UL2 \cite{tay2022ul2} with 19.5B parameters, Meta-Llama-3-70B-Instruct (AI@Meta, 2024), Mistral 7B \cite{jiang2023mistral}, Mixtral 7Bx8 \footnote{https://mistral.ai/news/mixtral-of-experts/}, and Claude 3.5 Sonnet \footnote{https://www.anthropic.com/news/claude-3-5-sonnet}. We directly download FLAN-T5-XXL and FLAN-UL2 models from Huggingface and run inference on a p4.24 xlarge EC2 instance with a batch size of 1. For other models, we run inference by making API calls. All inferences are performed with temperature=0.2 and top\_p=0.9. Principles are generated based on a sampling of n=[4, 8, 16] demonstrations with and without label information from the training set for each task. Refer to Appendix \ref{tab:irony2018} for prompt examples that we use to perform this step. Accordingly, for each task, we obtain 3 × 2 × 6 = 36 principle candidates by varying (1) the number of demonstrations: [4, 8, 16], (2) labeled or unlabeled demonstrations, and (3) six different LLM agents.

\subsubsection{Principle Consolidation}

After the \textbf{Principle Generation} step, we discard the analysis and extract the principles only. These 36 principle candidates are then sent to a finalizer agent to provide the optimal principle for performing the target classification task. We implement three methods based on the paradigm of how these principle candidates are utilized to derive the final principle:

(1) Listwise ranking by the finalizer agent: We directly ask each LLM to rank the top five principles given the entire list of candidate principles based on their helpfulness for performing the target classification task. Previous research shows that ICL is sensitive to permutation of in-context examples (i.e., selection and ordering) \cite{wu2022self}. Accordingly, we randomize the list of principles presented to LLMs in two different orders, with and without demonstrations (n=2) to illustrate how the target task is defined, yielding 2 × 2 = 4 different prompts for each LLM agent. We aggregate the top five ranked principles from each LLM agent and select the top 1 principle for each dataset based on majority voting. We use all LLMs mentioned above except FLAN-T5-XXL \cite{chung2024scaling} and FLAN-UL2 \cite{tay2022ul2} because they exceed the input token length limits of 512 or 2048 if we put all the candidate principles in one single prompt. This requires 4 × 4 = 16 inference costs from various multi-agent LLMs. See the Appendix for prompt examples for principle ranking and Table \ref{tab:dataset_principles} as an example of the final principle selected. 

(2) Consolidation by the finalizer agent: The listwise ranking method tries to make agents compete with each other and select the best principle based on their helpfulness to the downstream classification task. In contrast, the consolidation method acknowledges that a single agent might not be able to provide the optimal principle for the task and instead tries to establish a comprehensive principle by integrating and summarizing key points from all principles while resolving conflicting information. Since this method requires the LLM agent to possess reasoning capabilities, we select Claude 3.5 Sonnet as the finalizer agent based on the overall high quality of principles generated in the previous step. See Appendix for prompt examples for principle consolidation and Table \ref{tab:dataset_principles} as an example of the final principle selected. 

(3) Random selection (control group): This method randomly selects one principle from all the candidates.

\subsubsection{Text Classification}
After selecting the optimal principle for performing the downstream classification task, we append it to the prompt as context and ask LLMs to provide the answer to the classification task based on the provided principles. In this step, we only experiment with two open-source LLMs from Huggingface: FLAN-T5-XXL and FLAN-UL2, due to inference cost concerns. We run the inference with five random seeds on a p4.24xlarge EC2 instance using the same hyperparameters as in the \textbf{Principle Generation} step.

\subsection{Baselines}
We compare our PRINCIPLE-BASED PROMPTING approach to the following baselines. All prompting approaches listed here are considered single-agent approaches which involve one or multiple inferences with one LLM.
\subsubsection{Vanilla Prompting}
The LLM is provided with a task description containing all classification options, and then directly asked to provide a decision in short answer format. In the zero-shot setting, no demonstrations are provided, while n demonstrations are provided for few-shot settings.

\subsubsection{CoT Prompting}
The only difference between Vanilla and CoT prompting is that "Let's think step by step" is appended to prompts right before asking the LLM to output the final answer.

\subsubsection{Stepback Prompting}
In this two-step prompting approach, the LLM is first asked "What are the principles or important features to distinguish..." and then asked to provide the classification decision given the answers from the first step.

\subsubsection{Principle Single-Agent}
Unlike our multi-agent framework, this approach asks the classifier agent to first provide principles based on its analysis of randomly sampled demonstrations (n=4). Then these principles are appended as context when performing ICL for text classification tasks. We use this baseline to evaluate the contribution of the multi-agent framework to performance gains.

\subsubsection{Finetuning}
Finetuning RoBERTa-large in full or few-shot settings: We also finetune a pretrained language model (RoBERTa-large) on training data with a linear classification layer on top of [CLS] embeddings. For public datasets (Irony2018, Emotion20, and Financial), the training sets range from 1K to 4K samples. We also finetune RoBERTa-large with only 10\% of the datasets to evaluate performance in the few-shot settings. In contrast, two internal datasets PC1 and PC2 have very limited training data (\textless 200 samples), thus automatically falling into the few-shot setting.

\section{Experiments}
\subsection{Datasets}
We test our PRINCIPLE-BASED ICL approach and baselines on five text classification datasets: three are public (Irony2018, Emotion20 \cite{barbieri2020tweeteval}, \cite{sailunaz2019emotion} and Financial Phrasebank \cite{malo2014good}) and two are private datasets: Production Classification 1 (PC1) and Production Classification 2 (PC2). PC1, PC2, and Irony2018 are binary classification tasks, while Emotion20 and Financial Phrasebank are multi-class classification tasks.

\subsubsection{Irony2018}
We choose the Subtask 3A dataset of the SemEval2018 Irony Detection challenge \cite{barbieri2020tweeteval} (referred to as "Irony18"). The goal is to determine whether a tweet contains ironic intent. It contains 784 tweets in the test set.

\subsubsection{Emotion20}
Emotion recognition involves the identification and understanding of emotions expressed in text \cite{sailunaz2019emotion}. The objective of this dataset is to identify four emotions expressed: anger, joy, optimism, and sadness. We use the dataset provided by TweetEval benchmark \cite{barbieri2020tweeteval}, which we refer to as "Emotion20". It contains 1,421 data points in the test set.

\subsubsection{Financial Phrasebank}
We choose dataset of sentences labeled with polar sentiment from financial news. This dataset consists of 4,840 sentences from English-language financial news categorized by sentiment. It is divided by agreement rates of 5-8 annotators, and we select labels with instances having $\geq$75\% agreement. We refer to this dataset as "Financial". It contains 1,036 financial statements in the test set.

\subsubsection{PC1 and PC2}
Production Classification 1 and 2 are binary classification datasets consisting of product descriptions from an e-commerce website and their associated classes as labels. They contain 1,788 and 1,749 unique products in the test set, respectively.

\subsection{Evaluation}
We use the macro-averaged F1 score as the evaluation metric, which considers the overall performance across all classes.

\section{Results}
Table \ref{tab:zero-shot} shows that under zero-shot settings, our PRINCIPLE-BASED PROMPTING approach not only outperforms vanilla prompting but also other strong baselines such as CoT prompting and stepback prompting for both FLAN-T5-XXL and FLAN-UL2 models. The principle single-agent approach achieves on-par or better performance than the more costly stepback prompting approach. Stepback prompting incurs twice the inference costs of vanilla prompting due to its two-step prompting strategy at the instance level (one for eliciting abstracted principles via questions, one for classification decisions). In contrast, the principle single-agent approach only adds one single inference for generating principles at the task level. 

The multi-agent LLM framework with consolidation can further boost performance gains with the principle-based approach on top of single-agent implementation by 1.23\% on FLAN-T5-XXL and 6.52\% on FLAN-UL2 on average across five datasets. FLAN-UL2 with principles finalized by the multi-agent consolidation approach boosts model performance by 10.69\% over vanilla prompting averaged across five datasets. FLAN-T5-XXL also achieves 6.92\% performance gains averaged across five datasets. In general, the performance gains are more evident and consistent on FLAN-UL2 than FLAN-T5-XXL. This is probably due to FLAN-UL2's stronger reasoning capability with nearly twice as many model parameters as FLAN-T5-XXL, which can better incorporate principles provided to guide the downstream classification task. In comparison, other strong single-agent baselines such as CoT and stepback prompting either do not show consistent performance gains compared to vanilla prompting or are outperformed by the principle-based approach. For instance, FLAN-T5-XXL fails to benefit from CoT in general, while the multi-agent principle-based approach can further improve stepback prompting from 3.05\% to 6.92\% on FLAN-T5-XXL and from 4.28\% to 10.69\% on FLAN-UL2. 

Under the multi-agent framework, the consolidation approach performs better than its ranking and random (control group) counterparts. Interestingly, the ranking approach is sometimes even outperformed by random selection. This is likely because the cooperative mode of the multi-agent framework can better leverage different perspectives from multiple agents and potentially resolve limitations of single-agent approaches. In contrast, the competitive mode is too risky and more likely to fail, as it heavily relies on the capability of a single champion agent.

Additionally, both LLMs perform better on the PC2 private classification task using principles generated and finalized by the principle-based multi-agent consolidation approach compared to principles created by humans (16.21\% vs. 14.89\% on FLAN-T5-XXL and 19.37\% vs. 13.26\% on FLAN-UL2). This demonstrates the effectiveness of our approach. On PC1, the principle-based multi-agent ranking approach achieves comparable or better performance gains (1.57\% vs. 0.90\% on FLAN-T5-XXL and 3.71\% vs. 3.98\% on FLAN-UL2) compared to human-created principles.

When comparing with the finetuned RoBERTa-Large model, our PRINCIPLE-BASED PROMPTING approach significantly outperforms the finetuned encoder-only RoBERTa-Large under low-resource settings on three public datasets, using only 10\% of the labeled datasets, resulting in training sets ranging from 78 to 174 samples. Since PC1 and PC2 have fewer than 200 training samples, they automatically fall into few-shot settings. The advantages of supervised fine-tuning diminish under few-shot settings, showing negative performance gains compared to the zero-shot vanilla prompting approach across all five datasets. When the number of labeled data increases to the full dataset, which contains thousands of labeled samples, the finetuned RoBERTa-Large model's performance improves due to explicit supervision from these labels and finally outperforms LLM ICL approaches on Emotion20 (15.11\% vs. 17.62\%) and Financial (14.17\% vs. 16.62\%) by only small margins. The small performance gap demonstrates that our principle-based multi-agent LLM approach can serve as an effective and cost-friendly alternative to supervised classifiers when labeling resources are constrained.

\begin{table*}[htbp]
\centering
\caption{Absolute improvements in the macro-F1 scores over the zero-shot vanilla prompting for various single- and multi-agent approaches under the zero-shot settings. Human-crafted principles are only available for two private datasets. Results are averaged across five inferences with different random seeds.}

\label{tab:zero-shot}

\renewcommand{\arraystretch}{1.14}  

\begin{tabular}{p{1.4cm}|p{1.6cm}|p{3.3cm}|c|c|c|c|c|c} 
\hline
\toprule
    Model & \multicolumn{2}{c|}{Method}  & Irony2018 & Emotion20 & Financial & PC1 & PC2 & AVG \\
\midrule
\multirow{7}{*}{flan-t5-xxl} & 
\multirow{4}{*}{single agent} & 
    CoT & -9.31 & -14.23 & 1.51 & -1.56 & 17.25 & -1.27 \\

    & & stepback & -2.03 & 1.68 & -3.31 & 1.36 & 17.56 & 3.05 \\

    & & principle & 2.62 & 8.13 & 3.40 & 1.40 & 12.89 & 5.69 \\

    & & principle+human & NA & NA & NA & 3.98 & 14.89 & NA \\
\cline{2-9}
    & \multirow{3}{*}{multi agent} & 
    principle+random & 0.63 & 9.74 & 6.69 & 2.43 & 14.16 & 6.73 \\

    & & principle + ranking & 1.55 & 9.52 & 4.16 & \textbf{3.71} & 13.84 & 6.56 \\

    & & principle+consolidation & 0.45 & 12.13 & 4.38 & 1.43 & 16.21 & 6.92 \\
\midrule
\multirow{7}{*}{flan-ul2} & 
\multirow{4}{*}{single agent} & 
    CoT & -6.87 & 0.41 & 0.96 & -0.58 & 13.46 & 1.48 \\

    & & stepback & 2.72 & 0.47 & 4.18 & 0.02 & 13.99 & 4.28 \\

    & & principle & 4.57 & 0.02 & 3.42 & -0.2 & 13.03 & 4.17 \\

    & & principle + human & NA & NA & NA & 0.90 & 13.26 & NA \\
\cline{2-9}
    & \multirow{3}{*}{multi agent} & 
    principle+random & \textbf{5.56} & 12.15 & 11.78 & -0.54 & 19.08 & 9.61 \\

    & & principle+ranking & 4.96 & 11.14 & 11.05 & 1.57 & 18.69 & 9.48 \\

    & & principle+consolidation & 4.77 & \textbf{15.11} &\textbf{14.17} & 0.04 & \textbf{19.37} & \textbf{10.69} \\

\midrule
\multirow{2}{*}{RoBERTa} & full & \multirow{2}{*}{finetune} 
 &0.44 & \textbf{*17.62} & \textbf{*16.62} & -5.26 & -7.93 & 4.30 \\
 &10\% &  & -19.71 & -41.01& -52.41 & NA & NA & NA\\
\bottomrule
\hline
\end{tabular}

\end{table*}

\section{Principle-based vs. Few-shot ICL}
We also compare the performance of the multi-agent principle consolidation approach with the few-shot ICL approach. Results in Table \ref{tab:few_shot} align with findings in previous research that adding more demonstrations tends to improve LLM ICL performance across all datasets with both LLMs \cite{levy2022diverse}. However, we also observe that this effect quickly diminishes, and model performance plateaus and even decreases as n increases to 4 or 8. Table \ref{tab:few_shot} shows that the principle-based approach is very competitive even in comparison to the few-shot ICL, which leverages one or more demonstrations as contexts, thus resulting in significantly increased input token length. It outperforms all few-shot ICL (n=[1, 2, 4, 8]) on four (Irony2018, Emotion20, Financial, and PC2) out of five datasets with FLAN-UL2, and also shows comparable performance gains on PC1 (0.59 vs. 0.04). Although the results with FLAN-T5-XXL are slightly mixed, it outperforms all few-shot ICL (n=[1, 2, 4, 8]) on two (Emotion20 and Financial) out of five datasets and shows comparative performance to the best n-shot setting on Irony2018 (0.68 vs. 0.45), PC1 (1.49 vs. 1.43), and PC2 (17.36 vs. 16.21).

\begin{table*}[htbp]
\centering
\caption{Absolute improvements in the macro-F1 scores over the zero-shot vanilla prompting for the few-shot versus zero-shot principle-based approaches. Results are averaged across five inferences with different random seeds. n indicates the number of demonstrations per class. For PC1 and PC2, experiments were limited to $n \leq 2$  due to out-of-memory errors caused by long input token lengths.}
\label{tab:few_shot}
\begin{tabular}{l|l|c|c|c|c|c}
\hline
\toprule
Dataset & Model & n=1 & n=2 & n=4 & n=8 & \begin{tabular}[c]{@{}c@{}}multiagent\\principle consolidation\end{tabular} \\
\midrule
\multirow{2}{*}{irony2018} & flan-t5-xxl & 0.62 & 0.08 & 0.06 & \textbf{0.68} & 0.45 \\
& flan-ul2 & 3.63 & 3.08 & 3.64 & 3.66 & \textbf{4.77} \\
\midrule
\multirow{2}{*}{emotion20} & flan-t5-xxl & 7.82 & 4.17 & 1.92 & 2.58 & \textbf{12.13} \\
& flan-ul2 & 0.94 & 1.28 & 0.32 & 0.92 & \textbf{15.11} \\
\midrule
\multirow{2}{*}{financial} & flan-t5-xxl & 1.57 & 2.26 & 2.28 & 2.70 & \textbf{4.38} \\
& flan-ul2 & 8.22 & 10.42 & 11.49 & 11.32 & \textbf{14.17} \\
\midrule
\multirow{2}{*}{PC1} & flan-t5-xxl & 0.22 & \textbf{1.49} & NA & NA & 1.43 \\
& flan-ul2 & \textbf{0.59} & 0.47 & NA & NA & 0.04 \\
\midrule
\multirow{2}{*}{PC2} & flan-t5-xxl & \textbf{17.36} & 17.31 & NA & NA & 16.21 \\
& flan-ul2 & 16.98 & 17.41 & NA & NA & \textbf{19.37} \\
\bottomrule
\hline
\end{tabular}
\end{table*}

\begin{figure*}[htbp]
  \centering
  \includegraphics[scale=1]{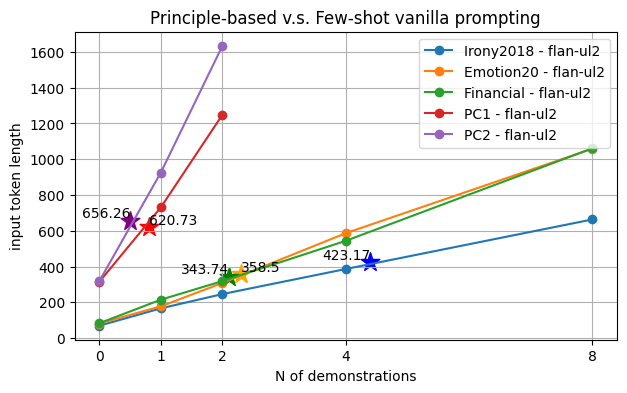} % Adjust width as needed
  \caption{Comparison of input token lengths between principle-based and few-shot vanilla prompting approaches. Stars on each line indicate where the input token length of the principle-based multi-agent consolidation approach corresponds to different n-shot settings (where n ranges from 1 to 8)}
  \label{fig:length}
\end{figure*}

Previous research adopts a sliding window approach to tackle the prompt length constraints \cite{ma2023zero, sun2023chatgpt} imposed by LLMs. We show our principle-based approach can also serve as a good solution to bypass this limit. We compute input token lengths of our multi-agent consolidation approach with both FLAN-UL2 and FLAN-T5-XXL tokenizers on each dataset. Since the numbers are very similar, we only use data from the FLAN-UL2 tokenizer. Figure \ref{fig:length} shows that the length of input tokens increases linearly as the number of demonstrations increases for few-shot prompting. In contrast, the principle-based approach has much shorter input token lengths compared to most few-shot settings. We can see that the input token length roughly corresponds to 2-shot on Emotion20 and Financial, and 4-shot on Irony 2018.

Since PC1 and PC2 are internal datasets with lengthy product titles and descriptions as inputs, increasing the number of demonstrations n beyond four in few-shot ICL is not only costly in terms of inference but also infeasible due to input length limits imposed by LLMs: 512 for FLAN-T5-XXL and 2048 for FLAN-UL2. The PRINCIPLE-BASED PROMPTING approach, however, only needs input token lengths that are even less than the 1-shot setting. Nevertheless, the PRINCIPLE-BASED PROMPTING approach achieves better performance on PC2 with FLAN-UL2 and comparable performance on PC1 with both models, while significantly reducing inference costs.

\section{Ablation Studies}
We further investigate how different factors contribute to crafting high-quality principles for predicting downstream classification tasks: (1) the number of demonstrations used for principle generation, (2) whether these demonstrations are labeled, and (3) the use of a single-agent versus multi-agent LLM framework. Specifically, we randomly sample demonstrations (where n=[4, 8, 16]) with and without labels. For the single-agent approach, we use the classifier LLM agent to generate principles based on its analysis of n demonstrations with or without label information. In the multi-agent approach, we employ the consolidation-based multi-agent LLM framework for each number of demonstrations. For both approaches, we use the same open-source models (FLAN-T5-XXL and FLAN-UL2) as classifier agents.

Figure \ref{fig:photo_label} shows that using more demonstrations does not guarantee higher quality principles during the principle generation stage. Including label information during principle generation, however, tends to have a positive impact on classification performance in most cases. Nevertheless, we observe exceptions on some datasets with different LLMs. For instance, FLAN-T5-XXL achieves better classification performance on Irony2018 and PC1 when using principles generated from unlabeled samples rather than labeled samples.

Additionally, as shown in Figure \ref{fig:photo_model}, principles generated by the multi-agent LLM framework significantly improve ICL performance across all datasets compared to those generated by the single-agent framework using relatively weaker LLMs (FLAN-T5-XXL and FLAN-UL2). This improvement is consistent across all numbers of demonstrations selected for principle generation, with the exception of 16-shot principle generation on Irony2018. These results demonstrate that our multi-agent consolidation framework is essential for generating high-quality principles for downstream classification. The framework overcomes the limitations of weaker classifier LLM agents (selected primarily due to inference cost considerations) by first utilizing LLMs with better reasoning capabilities (Claude 3.5 Sonnet and Llama-3-70B-Instruct) as principle generator agents, and then further optimizing principles through consolidation.

\section{Discussion and Conclusion}
We introduce PRINCIPLE-BASED PROMPTING, implemented via a multi-agent framework, as a simple yet generic strategy to elicit deep reasoning capabilities of LLMs by providing them with principles to perform downstream classification tasks. We show its superior performance over single-agent frameworks, including vanilla prompting and other strong ICL strategies such as CoT \cite{wei2022chain}, CARP \cite{sun2023text}, and stepback prompting \cite{zheng2023take}. One of the key differences between our work and previous works that attempt to scaffold LLMs with self-elicited clues or ask high-level concepts and principles before tackling the problem lies in our approach: instead of prompting LLMs to extract abstract principles or superficial clues to answer a single question, we perform knowledge distillation at the task level by providing multiple demonstrations with or without labels and instructing LLMs to extract common patterns (principles) based on their analysis. Our intuition is that analyzing how to solve the same task under different scenarios can help generate general knowledge that is abstracted away from details and thus easily applicable to unseen data with different distributions. The principles generated this way are knowledge-intensive and task-specific, and thus more efficient than those generated by purely relying on LLMs' general world knowledge obtained during the pretraining stage. 

Because principle generation is performed at the task level, we show that by implementing the principle-based approach via a multi-agent consolidation framework, we can achieve significant performance improvement with only minimal additional inference costs for text classification tasks.

The competitive performance of our principle-based approach compared to few-shot ICL settings indicates that naively adding more demonstrations is not an efficient way to teach LLMs the input-label mapping relationship on new tasks. On one hand, sub-optimal sampling of demonstrations might provide a biased perspective for tackling the task, thus becoming insufficient to perform well on more complex or challenging examples. On the other hand, adding more demonstrations can potentially introduce more noise, as the vast amount of details contained in demonstrations is not only challenging for LLMs to comprehend but also distracting, since some details might be irrelevant for performing the classification task at hand. Accordingly, performance could be negatively impacted, as we observe in Table \ref{tab:few_shot}. In contrast, our PRINCIPLE-BASED approach abstracts away all these irrelevant details based on analysis across multiple demonstrations and presents only the most salient instructions for LLMs to focus on. It can serve as an alternative to the popular few-shot ICL approach for performing classification tasks, especially when inference costs and input token length are constraints imposed by certain LLMs.
 
Additionally, our multi-agent framework for principle generation is generic and can be applied to any use cases that require synthetic text generation. It can automatically generate highly relevant and knowledge-intensive documents (e.g. SOPs) with only a handful of examples, regardless of availability of labeling resources. Although traditional Retrieval Augmented Generation (RAG) usually performs retrieval of relevant documents from existing data stores, our approach can automatically generate highly relevant documents or SOPs for any tasks. The comparable or even better classification performance of LLMs shown in Table \ref{tab:zero-shot} using principles that are LLM-generated in comparison to human-generated counterparts suggests a promising direction to automate SOP generation without compromising on the quality of SOPs generated. As future research, it would be also interesting to see how our PRINCIPLE-BASED approach can be integrated with RAG. 

While our principle-based approach provides an effective and efficient ICL solution for text classification under zero-shot settings, we acknowledge several limitations. First, it might not work well for classification problems with many labels since generating principles that cover all classes might lead to very lengthy content to be included as contexts. In this case, we could potentially generate principles for each class individually and use a retriever to fetch corresponding principles for top-k classes before performing downstream classification. Additionally, we only explore open-source models such as FLAN-T5-XXL and FLAN-UL2 as classifier agents due to inference cost constraints. In future work, we would like to investigate whether the same performance gains can be replicated with black-box LLMs such as GPT-4. Lastly, while we mainly focus on zero-shot settings of our principle-based approach, it would also be interesting to explore whether adding concrete examples that are specifically analyzed and explained based on these principles would further improve model performance. We leave these research questions for future work.

\bibliography{principle}

% Appendix section
\appendix
\section{Appendix}

\begin{figure*}[htbp]
  \centering
  \includegraphics[width=1\textwidth]{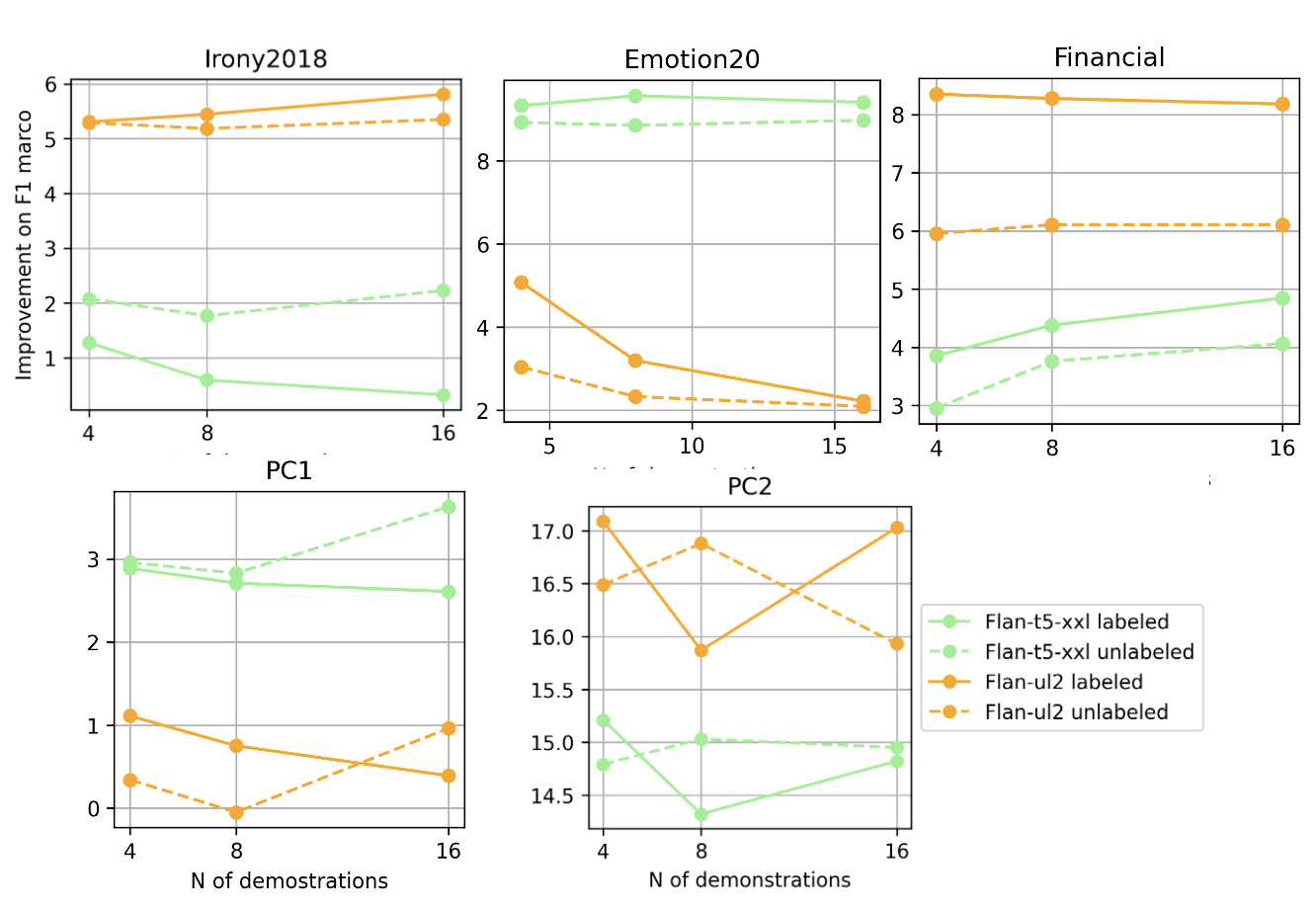} % Adjust width as needed
  \caption{Effects of label information in sampled demonstrations on generating high-quality principles for downstream classification}
  \label{fig:photo_label}
\end{figure*}

\begin{figure*}[htbp]
  \centering
  \includegraphics[width=1\textwidth]{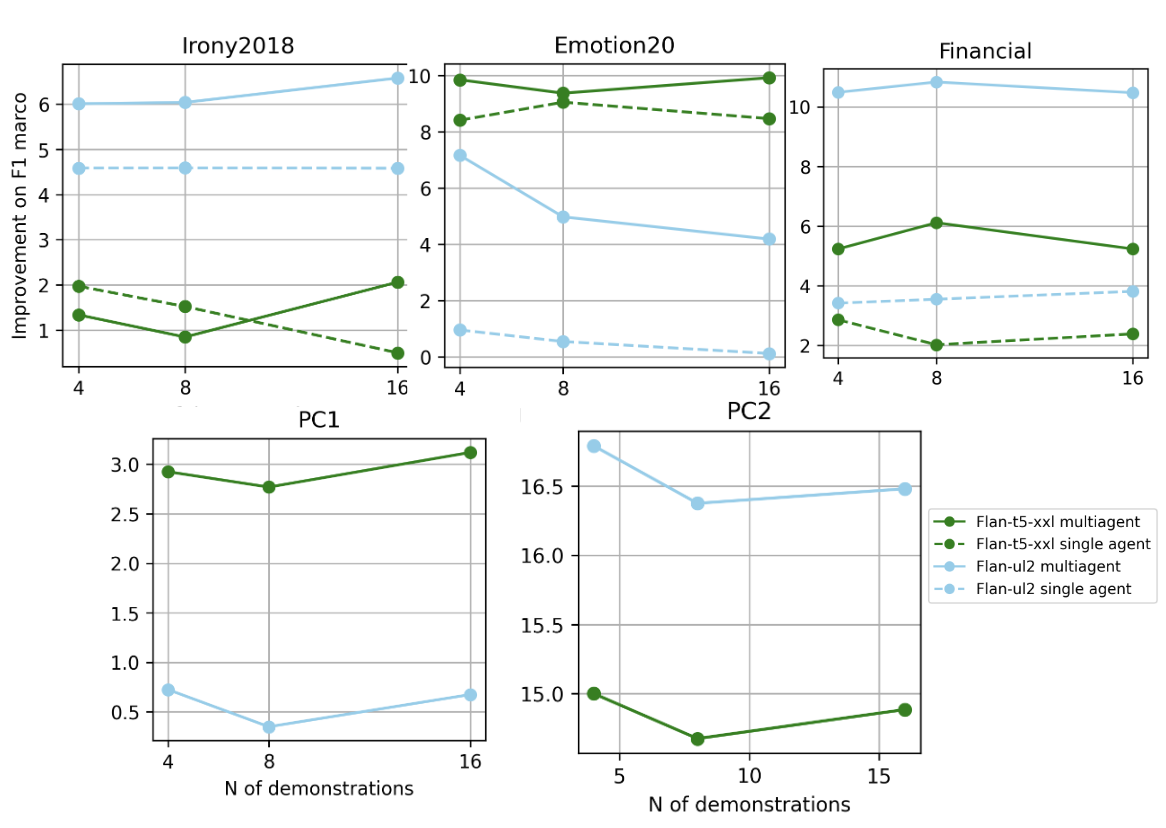} % Adjust width as needed
  \caption{Effects of single vs multi-agent in generating high-quality principles for downstream classification task}
  \label{fig:photo_model}
\end{figure*}

\begin{table*}[htbp]
    \centering
    \caption{Irony 2018}
    \label{tab:irony2018}
    \begin{tabular}{|l|p{0.6\textwidth}|}
        \toprule
        \textbf{Field} & \textbf{Description} \\
        \midrule
        \multirow{2}{*}{Label Word Mapping} & \{Yes: 1; No: 0\} \\
        \midrule
        \multirow{4}{*}{Principle Generation Prompt} & You are given the task to extract principles or important features which distinguish between statements that contain irony and those that do not. \\
        & Here are some examples: \\
        & Statement: $<$sent$>$ \\
        & Statement: $<$sent$>$ \\
        & Statement: $<$sent$>$ \\
        & Statement: $<$sent$>$ \\
        & Can you analyze each statement and identify whether it contains irony or not? \\
        & Based on your analysis, can you extract principles or important features which distinguish between statements that contain irony and those that do not? \\
        \midrule
        \multirow{3}{*}{Classification Prompt} & You are given the task to identify the sentiment of the following statement. \\
        & Here are important features to distinguish statements that contain irony and those that do not. \\
        & \{principle\} \\
        \midrule
        \multirow{3}{*}{Listwise Ranking Prompt} & You are given the task to rank a list of principles based on how helpful they are for identifying whether statements contain irony or not. \\
        & Here is the list of principles: \\
        & \{list of principles\} \\
        & Here are some examples of statements: \\
        & \{few\_shot\_example\} \\
        & How would you rank the principles above based on helpfulness for identifying whether statements contain irony or not? \\
        & Provide your ranking of top 10 principles in the following format: $A > B > C$... \\
        \midrule
        \multirow{3}{*}{Consolidation Prompt} & You are given multiple sets of principles for distinguishing emotions in statements. Your task is to analyze these principles and consolidate them into a single, comprehensive set of principles. \\
        & Here are the sets of principles: \\
        & \{sets of principles\} \\
        & Please analyze these principles and create a consolidated set that captures the most important and effective principles for identifying emotions in statements. Ensure the consolidated set is clear, non-redundant, and comprehensive. \\
        \bottomrule
    \end{tabular}
\end{table*}

\begin{table*}[htbp]
    \centering
    \caption{Emotion20}
    \label{tab:emotion20}
    \begin{tabular}{|l|p{0.6\textwidth}|}
        \toprule
        \textbf{Field} & \textbf{Description} \\
        \midrule
        \multirow{2}{*}{Label Word Mapping} & \{Anger: 0; Joy: 1; Optimism: 2; Sadness 3\} \\
        \midrule
        \multirow{4}{*}{Principle Generation Prompt} & You are given the task to extract principles or important features which distinguish statements that express four different emotions: anger, joy, optimism, and sadness. \\
        & Here are some examples that express different emotions: \\
        & Statement: $<$sent$>$ \\
        & Statement: $<$sent$>$ \\
        & Statement: $<$sent$>$ \\
        & Statement: $<$sent$>$ \\
        & Can you analyze each statement and identify the emotion that it tries to express from these four options: anger, joy, optimism, and sadness? \\
        & Based on your analysis, can you extract principles or important features which distinguish between statements that express these four emotions: anger, joy, optimism, and sadness? \\
        \midrule
        \multirow{3}{*}{Classification Prompt} & You are given the task to identify the emotion of the following statements from four options: anger, joy, optimism, and sadness. \\
        & Here are some principles that distinguish statements expressing different emotions: \\
        & \{principle\} \\
        \midrule
        \multirow{3}{*}{Listwise Ranking Prompt} & You are given the task to rank a list of principles based on how helpful they are for identifying the emotions of statements from four options: anger, joy, optimism, and sadness. \\
        & Here is the list of principles: \\
        & \{list of principles\} \\
        & Here are some examples of statements: \\
        & \{few\_shot\_example\} \\
        & How would you rank the principles above based on helpfulness for identifying emotions of statements? \\
        & Provide your ranking of top 5 principles in the following format: $A > B > C$... \\
        \midrule
        \multirow{3}{*}{Consolidation Prompt} & You are given a list of principles written by different LLM agents to distinguish statements that express four different emotions: anger, joy, optimism, and sadness. \\
        & Here are the sets of principles: \\
        & \{sets of principles\} \\
        & Please analyze these principles and create a consolidated set that captures the most important and effective principles for identifying irony in statements. Ensure the consolidated set is clear, non-redundant, and comprehensive. \\

        \bottomrule
    \end{tabular}
\end{table*}

\begin{table*}[htbp]
    \centering
    \caption{Financial}
    \label{tab:financial}
    \begin{tabular}{|l|p{0.6\textwidth}|}
        \toprule
        \textbf{Field} & \textbf{Description} \\
        \midrule
        \multirow{2}{*}{Label Word Mapping} & \{Positive: 1; Negative: 0; Neutral 2\} \\
        \midrule
        \multirow{4}{*}{Principle Generation Prompt} & You are given the task to extract principles or important features which distinguish between financial news that have positive, neutral, or negative sentiments. \\
        & Here are some examples: \\
        & Statement: $<$sent$>$ \\
        & Statement: $<$sent$>$ \\
        & Statement: $<$sent$>$ \\
        & Statement: $<$sent$>$ \\
        & Can you analyze each financial news below and identify the sentiment from these three options? \\
        & Based on your analysis, can you extract principles or important features which distinguish between statements that have positive, neutral, or negative sentiments? \\
        \midrule
        \multirow{3}{*}{Classification Prompt} & You are given the task to identify the sentiment of the following financial news. \\
        & Here are some key principles that distinguish statements with positive, neutral, and negative sentiments. \\
        & \{principle\} \\
        \midrule
        \multirow{3}{*}{Listwise Ranking Prompt} & You are given the task to rank a list of principles based on how helpful they are for identifying sentiments of financial news from three options: positive, negative, or neutral. \\
        & Here is the list of principles: \\
        & \{list of principles\} \\
        & Here are some examples of statements: \\
        & \{few\_shot\_example\} \\
        & How would you rank the principles above based on helpfulness for identifying sentiments of financial news? \\
        & Provide your ranking of top 10 principles in the following format: $A > B > C$... \\
        \midrule
        \multirow{3}{*}{Consolidation Prompt} & You are given a list of principles written by different LLM agents to distinguish financial news with positive, neutral or negative sentiments. \\
        & Here are the sets of principles: \\
        & \{sets of principles\} \\
        & Please analyze these principles and create a consolidated set that captures the most important and effective principles for identifying different sentiments in financial news. Ensure the consolidated set is clear, non-redundant, and comprehensive. \\
        \bottomrule
    \end{tabular}
\end{table*}

\begin{table*}[htbp]
    \centering
    \caption{PC1 and PC2}
    \label{tab:general_toys}
    \begin{tabular}{|l|p{0.6\textwidth}|}
        \toprule
        \textbf{Field} & \textbf{Description} \\
        \midrule
        \multirow{2}{*}{Label Word Mapping} & \{Yes: 1; No: 0\} \\
        \midrule
        \multirow{4}{*}{Principle Generation Prompt} & You are given the task to extract principles or important features which distinguish between products that are classified as A and those that are not. \\
        & Here are some examples and their corresponding answers. \\
        & Statement: $<$sent$>$ \\
        & Statement: $<$sent$>$ \\
        & Statement: $<$sent$>$ \\
        & Statement: $<$sent$>$ \\
        & Can you analyze each product description below and identify whether it is classified as A or not? \\
        & Based on your analysis, can you extract principles or important features which distinguish between products that are classified as A and those that are not? \\
        \midrule
        \multirow{3}{*}{Classification Prompt} & You are given the task to identify whether the product below is classified as A or not based on the product description. \\
        & Here are some key principles that distinguish products that are classified as A and those that are not. \\
        & \{principle\} \\
        \midrule
        \multirow{3}{*}{Listwise Ranking Prompt} & You are given the task to rank a list of principles based on how helpful they are for identifying whether products below are classified as A or not based on product descriptions. \\
        & Here is the list of principles: \\
        & \{list of principles\} \\
        & Here are some examples of statements: \\
        & \{few\_shot\_example\} \\
        & How would you rank the principles above based on helpfulness for identifying products as A or not? \\
        & Provide your ranking of top 5 principles in the following format: $A > B > C$... \\
        \midrule
        \multirow{3}{*}{Consolidation Prompt} & You are given a list of principles written by different LLM agents to distinguish products that are classified as A or not. \\
        & Here are the sets of principles: \\
        & \{sets of principles\} \\
        & Please analyze these principles and create a consolidated set that captures the most important and effective principles for identifying products classified as A or not. Ensure the consolidated set is clear, non-redundant, and comprehensive. \\
        \bottomrule
    \end{tabular}
\end{table*}

\begin{table*}[htbp]
    \centering
    \caption{ Principles examples finalized by multi-agent LLM framework}
    \label{tab:dataset_principles}
    \begin{tabular}{|p{1.5cm}|p{10cm}|}
        \toprule
        \textbf{Dataset} & \textbf{Principles finalized} \\
        \midrule
        Emotion20 & 
        Here are some principles that distinguish statements expressing different emotions:
        \begin{itemize}
            \item Anger statements tended to express resentment, insults, confrontation, aggression or rage. They often involved critique of others or expressed a desire for revenge.
            \item Joy statements conveyed a sense of cheerfulness, amusement or pleasure. They referenced positive or fun activities and did not criticize others. 
            \item Optimism statements had an upbeat, hopeful or ambitious tone. They focused on positive goals, beliefs in achievement or maintaining a positive mindset.
            \item Sadness statements expressed regret, disheartenment, grief, failure or negative outcomes. They had a somber, downbeat tone and referenced disappointment or undesirable situations.
        \end{itemize}
        Some key distinguishing features between the emotions included:
        \begin{itemize}
            \item Tone (positive vs. negative, upbeat vs. downbeat)
            \item Attitude toward others (critical vs. not critical) 
            \item Focus (goals/beliefs vs. regret/failure)
            \item References to emotion words like rage, disgust, cheerfulness, hope, regret
            \item Mention of confrontation/aggression vs. pleasure/amusement 
            \item Desire for revenge/payback vs. absence of such sentiments.
        \end{itemize} \\
        \midrule
        Irony2018 & 
        Key principles that distinguish statements that contain irony and those do not:
        \begin{itemize}
            \item - Ironic statements often use exaggerated, insincere or inappropriate language that implies the opposite or a hidden meaning when taken literally.
            \item - Ironic statements commonly employ linguistic cues like sarcasm, understatement or rhetorical questions to imply the unstated attitude of the speaker. 
            \item - Emoticons, punctuation or contextual cues can indicate a statement is not meant to be taken at face value.
            \item - Non-ironic statements directly and literally state what is meant without implicit, implied or hidden meanings beneath the surface. They do not rely on tone or context.
            In summary, ironic statements tend to have layers of implied or intended meaning beyond the surface interpretation, while non-ironic statements clearly and directly state what is meant without implicit meanings or implications. The use of exaggerated language, insincere tones and cues from context/punctuation also distinguishes ironic statements.

        \end{itemize} \\
        \bottomrule
    \end{tabular}
\end{table*}

\end{document}